\documentclass[10pt,journal,compsoc]{IEEEtran}

\ifCLASSOPTIONcompsoc
  \usepackage[nocompress]{cite}
\else
  \usepackage{cite}
\fi

\usepackage[utf8]{inputenc} % allow utf-8 input
\usepackage[T1]{fontenc}    % use 8-bit T1 fonts
\usepackage{hyperref}       % hyperlinks
\usepackage{url}            % simple URL typesetting
\usepackage{booktabs}       % professional-quality tables
\usepackage{amsfonts}       % blackboard math symbols
\usepackage{nicefrac}       % compact symbols for 1/2, etc.
\usepackage{microtype}      % microtypography
\usepackage{xcolor}         % colors
\usepackage{amsmath}
\usepackage{xspace}
\usepackage{amsfonts}
\usepackage{graphicx}
\usepackage{float}
\usepackage{courier}
\usepackage{wrapfig}
\usepackage{algorithm}
\usepackage{algorithmic}
\usepackage[figuresright]{rotating}

\usepackage{listings}
\usepackage{color}

\newcommand{\gmh}[1]{\textcolor[rgb]{0, 0, 0}{#1}}
\newcommand{\lzn}[1]{\textcolor[rgb]{0, 0, 0}{#1}}
\newcommand{\mtj}[1]{\textcolor[rgb]{0, 0, 0}{#1}}

\newcommand{\tabincell}[2]{\begin{tabular}{@{}#1@{}}#2\end{tabular}}

\newfloat{figtab}{htb}{fgtb}
\makeatletter
  \newcommand\figcaption{\def\@captype{figure}\caption}
  \newcommand\tabcaption{\def\@captype{table}\caption}
\makeatother

\definecolor{dkgreen}{rgb}{0,0.6,0}
\definecolor{gray}{rgb}{0.5,0.5,0.5}
\definecolor{mauve}{rgb}{0.58,0,0.82}

\ifCLASSINFOpdf
\else
\fi

\begin{document}
\title{Beyond Self-attention:  External Attention using Two Linear Layers for Visual Tasks}

\author{Meng-Hao Guo,
        Zheng-Ning Liu,
        Tai-Jiang Mu,
        Shi-Min Hu,~\IEEEmembership{Senior Member,~IEEE,}% <-this % stops a space

\IEEEcompsocitemizethanks{\IEEEcompsocthanksitem M.H.~Guo, Z.N.~Liu, T.J.~Mu and S.M.~Hu are with the BNRist, Department of Computer Science and Technology, Tsinghua University, Beijing 100084, China.%\protect\\
% % note need leading \protect in front of \\ to get a newline within \thanks as
% % \\ is fragile and will error, could use \hfil\break instead.
%E-mail: see http://www.michaelshell.org/contact.html
\IEEEcompsocthanksitem S.M.~Hu is the corresponding author.\protect\\ 
E-mail: shimin@tsinghua.edu.cn.}% <-this % stops an unwanted space
}


\markboth{Journal of \LaTeX\ Class Files,~Vol.~14, No.~8, August~2015}%
{Shell \MakeLowercase{\textit{et al.}}: Bare Demo of IEEEtran.cls for Computer Society Journals}



\IEEEtitleabstractindextext{%
% !TEX root = ../main.tex
\begin{abstract} 
Attention mechanisms, especially self-attention, have played an increasingly important role in deep feature representation for visual tasks. 
Self-attention updates the feature at each position by computing a weighted sum of features using pair-wise affinities across all positions to capture the long-range dependency within a single sample. 
However, self-attention has quadratic complexity and ignores potential correlation between different samples.
This paper proposes a novel attention mechanism which we call \emph{external attention}, based on two external, small, learnable, shared memories, which can be implemented easily by simply using two cascaded linear layers and two normalization layers; it conveniently replaces self-attention in existing popular architectures. 
External attention has linear complexity and implicitly considers the correlations between all \lzn{data} samples.
\lzn{We further incorporate the multi-head mechanism into external attention to %capture multiple attentions in parallel. Finally, we 
provide an all-MLP architecture, \emph{external attention MLP} (EAMLP), for image classification.} %based on the proposed multi-head external attention. }
% \gmh{We also design a multi-head external attention. Benefit from the proposed multi-head external attention, we achieve an all MLPs architecture for image recognition.}
%can conveniently replace self-attention in the popular architectures such as DANet, SAGAN and T2T-Transformer. 
% Extensive experiments on image classification, \gmh{object detection}, semantic segmentation, \gmh{instance segmentation}, image generation, point cloud classification and point cloud segmentation tasks 
\lzn{Extensive experiments on image classification, object detection, semantic segmentation, instance segmentation, image generation, and point cloud analysis}
reveal that our method provides results comparable or superior   to the self-attention mechanism and some of its variants, with much lower computational and memory costs. 
\end{abstract}

\begin{IEEEkeywords}
Deep Learning, Computer Vision, Attention, Transformer, Multi-Layer Perceptrons.
\end{IEEEkeywords}
}

\maketitle

\IEEEdisplaynontitleabstractindextext

% \fi
\IEEEpeerreviewmaketitle

% \cite{streater2016pct}
% !TEX root = ../main.tex
\IEEEraisesectionheading{\section{Introduction}\label{sec:introduction}}

\IEEEPARstart{D}{ue} to its ability to capture long-range dependencies, the self-attention mechanism helps to improve performance in various natural language processing~\cite{Bahdanau2015neural,Devlin2019bert} and computer vision~\cite{Wang_2018_CVPR, fu2019dual} tasks. 
Self-attention works by refining the representation at each position via aggregating features from all other locations in a single sample,  
%However, since self-attention needs to aggregate feature from all other locations 
which leads to quadratic computational complexity in the number of locations in \gmh{a} sample. 
Thus, some variants attempt to approximate self-attention at a lower computational cost~\cite{huang2018ccnet,li19,yuan2020objectcontextual,geng2021is}. 
%Huang et al.~\cite{huang2018ccnet} propose criss-cross attention, which first adopts row attention and then uses column attention to capture the global context. Li et al.~\cite{li19} propose Expectation-Maximization(EM) attention mechanism, applying Expectation-Maximization clustering to optimize self-attention. Yuan et al.~\cite{yuan2020objectcontextual} propose to use object-contextual vector as center to approximate attention. Geng et al.~\cite{geng2021is} show that matrix decomposition is a better way to replace self-attention. 

Furthermore, self-attention concentrates on the self-affinities between different locations within a single sample, and ignores potential correlations with other samples. 
%To the best of our knowledge, this is the first attempt to solve this problem in visual tasks. 
It is easy to see that incorporating correlations between different samples can help to contribute to a better feature representation. 
For instance,  features belonging to the same category but distributed across different samples should be treated consistently in \gmh{the} semantic segmentation task, and a similar observation applies in image classification and various other visual tasks.%~\cite{}.

This paper proposes a novel lightweight attention mechanism which we call  \emph{external attention} (see Figure~\ref{fig:attention}c)). 
As shown in Figure~\ref{fig:attention}a), computing self-attention requires  first calculating an attention map by computing the affinities between self query vectors and self key vectors, then generating a new feature map by weighting the self value vectors with this attention map. 
External attention works differently. We first calculate the attention map by computing the affinities between the self query vectors and an external learnable \emph{key} memory, and then produce a refined feature map by multiplying this attention map by another external learnable \emph{value} memory. 

In practice, the two memories are implemented with linear layers, and can thus be optimized by back-propagation in an end-to-end manner. 
They are independent of individual samples and shared across the entire dataset, which plays a strong regularization role and improves the generalization capability of the attention mechanism. 
The key to the lightweight nature of external attention is that the number of elements in the memories is much smaller than the number in the input feature, yielding a  computational complexity linear in the number of elements in the input. 
The external memories are designed to learn the most discriminative features across the whole dataset, capturing the most informative parts, as well as excluding interfering information from other samples.
A similar idea can be found in sparse coding~\cite{sc-nature96} or dictionary learning~\cite{DicLearning06}. 
Unlike those methods, however, we neither try to reconstruct the input features nor apply any explicit sparse regularization to the attention map.

Although the proposed external attention approach is simple, it is effective for various visual tasks. Due to its simplicity, it can be easily incorporated into existing popular self-attention based architectures, such as DANet~\cite{fu2019dual}, SAGAN~\cite{Zhang2019sagan} and T2T-Transformer~\cite{yuan2021tokenstotoken}.
Figure~\ref{fig:seg_aritecture} demonstrates a typical 
% encoder-decoder like 
architecture replacing self-attention with our external attention for an image semantic segmentation task.
We have conducted extensive experiments on such basic visual tasks as classification, object detection, semantic segmentation, and instance segmentation and generation, with different input modalities (images and point clouds).% classification,  semantic  segmentation,  image  generation,  point cloud classification and point cloud segmentation 
The results reveal that our method achieves results comparable to or better  than the original self-attention  mechanism  and  some  of  its  variants, at much lower computational cost.

\lzn{To learn different aspects of attentions for the same input, we incorporate the multi-head mechanism into external attention, boosting its capability. 
\mtj{Benefiting from the proposed multi-head external attention,}  
%Furthermore, 
we have designed a novel all-MLP architecture named EAMLP, which \mtj{is comparable to CNNs and the original Transformers for the image classification task.} %is benefit from the proposed multi-head external attention. Compared to CNNs and Transformers, the proposed all-MLP architecture is hardware-friendly and easy to optimize.
}
% the proposed multi-head mechanism does not increase computation complexity, unlike multi-head self-attention. 

\gmh{The main contributions of this paper are summarized below:}

\begin{itemize}
\item\gmh{ A novel attention mechanism, external attention, with $O(n)$ complexity; \mtj{it can replace self-attention in existing architectures}. It can mine potential relationships across the whole dataset, affording a  strongly regularizing role, and improving the generalization capability of the attention mechanism. }

\item \gmh{ Multi-head external attention, which benefits us to build an all MLP architecture; it achieves a  top1 accuracy of 79.4\% on the ImageNet-1K dataset.}

\item \gmh{Extensive  experiments utilizing external attention for
image classification, object detection, semantic  segmentation, instance segmentation, image  generation, point cloud classification, and point cloud segmentation. In scenarios where computational effort must be kept low, it achieves better results than the original self-attention  mechanism  and some  of  its  variants.}

\end{itemize}

% !TEX root = ../main.tex

\section{Related work}
Since a comprehensive review of the attention mechanism is beyond the scope of this paper, we only discuss the most closely related literature in the vision realm.

\subsection{The attention mechanism in visual tasks}

The attention mechanism can be viewed as a mechanism for reallocating resources according to the importance of activation. It plays an important role in the human visual system. There has been vigorous development of this field in the last decade~\cite{mnih2014recurrent, xu2016show, Hu2018SENET, Vaswani2017attention, Wang_2018_CVPR, yuan2019ocnet, Dosovitskiy2020ViT}. 
Hu et al. proposed SENet~\cite{Hu2018SENET}, showing that the attention mechanism can reduce noise and improve  classification performance. 
Subsequently, many other papers have applied it to visual tasks. 
Wang et al. presented non-local networks~\cite{Wang_2018_CVPR} for video understanding,
Hu et al.~\cite{hu2018relation} used attention in object detection, 
Fu et al. proposed DANet~\cite{fu2019dual} for semantic segmentation, 
Zhang et al.~\cite{Zhang2019sagan} demonstrated the effectiveness of the attention mechanism in image generation, and  
Xie et al. proposed A-SCN~\cite{Xie_2018_CVPR} for point cloud processing.

\subsection{Self-attention in visual tasks}
Self-attention is a special case of attention, and many papers~\cite{Wang_2018_CVPR, yuan2019ocnet, fu2019dual, Zhang2019sagan, aanet}, have considered the self-attention mechanism for vision. The core idea of self-attention is calculating the affinity between features to capture  long-range dependencies. 
However, as the size of the feature map increases, the computing and memory overheads increase quadratically. 
To reduce computational and memory costs, Huang et al.~\cite{huang2018ccnet} proposed criss-cross attention, which considers row attention and  column attention in turn to capture the global context. 
Li et al.~\cite{li19} adopted expectation maximization (EM) clustering to optimize self-attention. 
Yuan et al.~\cite{yuan2020objectcontextual} proposed use of object-contextual vectors to process attention; however, it depends on semantic labels. 
Geng et al.~\cite{geng2021is} show that matrix decomposition is a better way to model the global context in semantic segmentation and image generation. Other works~\cite{ramachandran2019standalone, san_zhao} also explore extracting local information by using the self-attention mechanism.

Unlike self-attention which obtains an attention map by computing affinities between self queries and self keys, our external attention computes the relation between self queries and a much smaller learnable key memory, which captures the global context of the dataset. 
External attention does not rely on semantic information and can be optimized by the back-propagation algorithm in an end-to-end way instead of requiring an iterative algorithm.

\subsection{Transformer in visual tasks}

Transformer-based models have had great success in natural language processing~\cite{Bahdanau2015neural, Lin2017selfatt, Vaswani2017attention, Devlin2019bert, Yang2019xlnet, Dai2019txl, Lee2020biobert}. Recently, they have also  demonstrated huge potential for visual tasks. 
Carion et al.~\cite{Carion2020e2e} presented an end-to-end detection transformer that takes CNN features as input and generates bounding boxes with a transformer.
Dosovitskiy~\cite{Dosovitskiy2020ViT} proposed ViT, based on patch encoding and a transformer, showing that with sufficient training data, a transformer provides better performance than a traditional CNN. 
Chen et al.~\cite{chen2020generative} proposed iGPT for image generation based on use of a transformer.

Subsequently, transformer methods have been successfully applied to many visual tasks, including %  DeiT~\cite{touvron2021training},  T2T-ViT~\cite{yuan2021tokenstotoken},  for 
image classification~\cite{touvron2021training,yuan2021tokenstotoken, liu2021swin, fan2021multiscale}, %Deformable DETR~\cite{zhu2021deformable} for 
object detection~\cite{zhu2021deformable}, %IPT~\cite{chen2020pretrained} for 
lower-level vision~\cite{chen2020pretrained},  %SETR~\cite{zheng2020rethinking} for
semantic segmentation~\cite{zheng2020rethinking}, %TransT~\cite{TransT} for 
tracking~\cite{TransT}, % VisTR~\cite{wang2020endtoend} for 
video instance segmentation~\cite{wang2020endtoend},  %TransGAN~\cite{jiang2021transgan} for 
image generation~\cite{jiang2021transgan}, %UniT~\cite{hu2021transformer} for 
multimodal learning~\cite{hu2021transformer}, %TransReID~\cite{he2021transreid} for 
object re-identification~\cite{he2021transreid}, %CPTR~\cite{liu2021cptr} for
image captioning~\cite{liu2021cptr}, %PCT~\cite{guo2020pct} for 
point cloud learning~\cite{guo2021pct} and self-supervised learning~\cite{chen2021empirical}. 
Readers are referred to recent surveys~\cite{han2021survey, khan2021transformers} for a more comprehensive review of the use of transformer methods for visual tasks.

\begin{figure}[t]
    \centering
    \includegraphics[width=\linewidth]{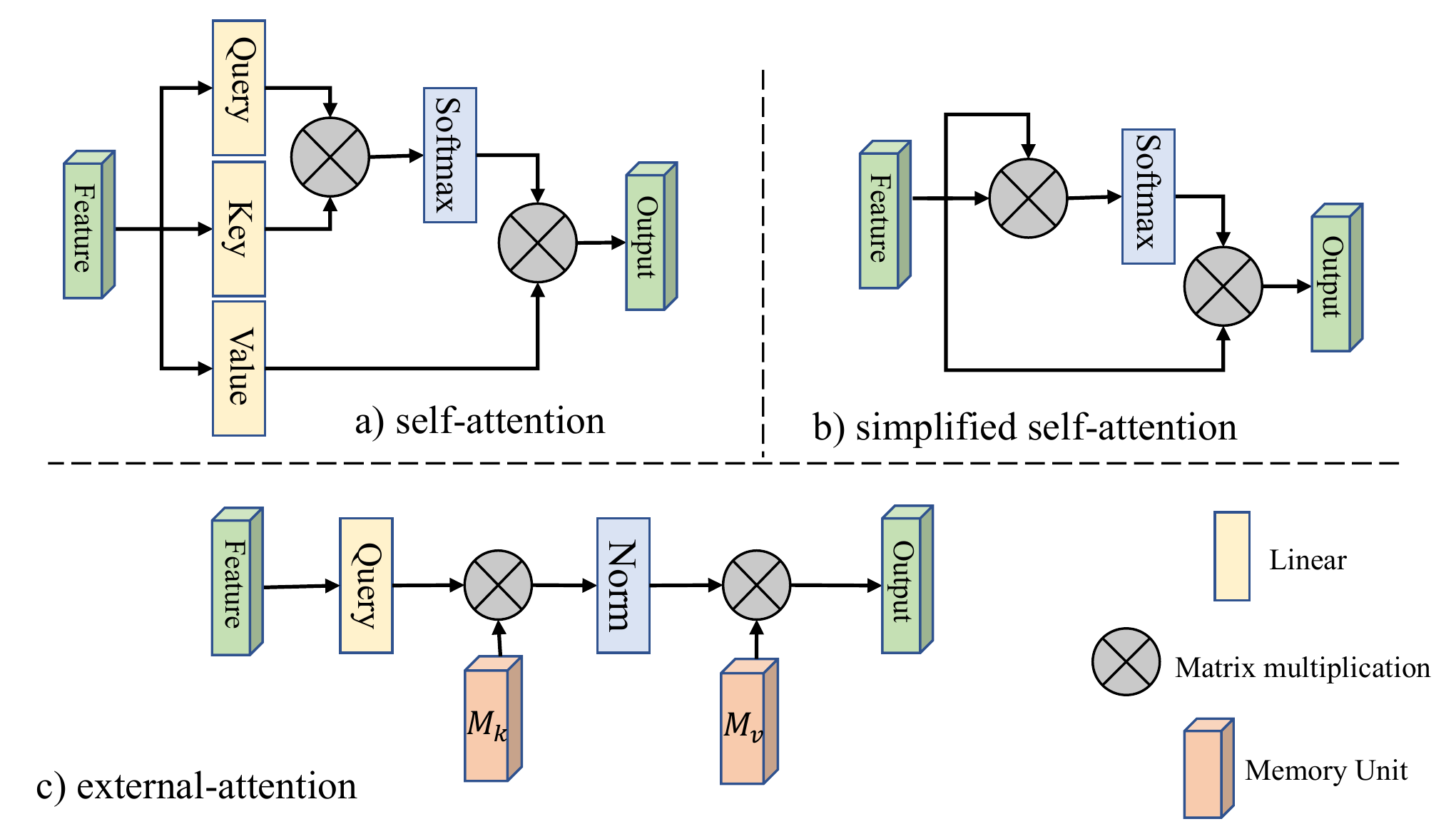}
    \caption{
    Self-attention versus external-attention
    }
    \label{fig:attention}
    %\vspace{-3ex}
\end{figure}

% !TEX root = ../main.tex

\section{Methodology}

In this section, we start by analyzing the original self-attention mechanism.
% and point out it shortcomings. 
Then we detail our novel way to define attention: external attention. It can be implemented easily by only using two linear layers and two normalization layers, as later shown in Algorithm~\ref{alg.single_head}. 
%Finally, the complete architecture of EANet will be shown. 

% \subsection{Replace self-attention by two linear layers}

\subsection{Self-Attention and External Attention}
We first revisit the self-attention mechanism (see Figure~\ref{fig:attention}a)). 
Given an input feature map $F \in \mathbb{R}^{N \times d}$, where $N$ is the number of \mtj{elements (or }pixels \mtj{in images)} and $d$ is the number of feature dimensions, self-attention linearly projects the input to a query matrix $Q \in \mathbb{R}^{N \times d'}$, a key matrix $K \in \mathbb{R}^{N \times d'}$, and a value matrix $V \in \mathbb{R}^{N \times d}$ ~\cite{Vaswani2017attention}. Then self-attention can be formulated as:
\begin{align}
    A & = (\alpha)_{i,j} = \mathrm{softmax}(Q  K^T), \\
    F_{out} & = A V, 
\end{align}
where $A \in \mathbb{R}^{N \times N} $ is the attention matrix and $\alpha_{i,j}$ is the pair-wise affinity  between (similarity of) the $i$-th and $j$-th elements. 

A common simplified variation (Figure~\ref{fig:attention}b)) of self-attention directly calculates an attention map from the input feature $F$ using:
\begin{align}
    A & = \mathrm{softmax}(F F^T), \\
    F_{out} & = A  F.
\end{align}
Here, the attention map is obtained by computing pixel-wise similarity in the feature space, and the output is the refined feature representation of the input.

However, even when simplified, the high computational complexity of ${O}(dN^2)$ presents a significant drawback to use of self-attention. The quadratic complexity in the number of input pixels makes direct application of self-attention to images infeasible. Therefore, previous work~\cite{Dosovitskiy2020ViT} utilizes self-attention on patches rather than pixels to reduce the computational effort.

\gmh{Self-attention can be viewed as using a linear combination of self values to refine the input feature. However, it is far from obvious that we really need $N\times N$ self attention matrix and an $N$ element self value matrix in this linear combination. Furthermore, self-attention only considers the relation between elements within a data sample and ignores potential relationships between elements in different samples, potentially limiting the ability and flexibility of self-attention.}

Thus, we propose \gmh{a novel attention module named} \emph{external attention}, which computes attention between the input pixels and an external memory unit $M \in \mathbb{R}^{S \times d}$, via:
\begin{align}
\label{eq:ea}
    A & = (\alpha)_{i,j} = \mathrm{Norm}(F M^T), \\
    F_{out} & = A  M.
\end{align}

Unlike self-attention, $\alpha_{i,j}$ in Equation~(\ref{eq:ea}) is the similarity between the $i$-th pixel and the $j$-th row of $M$, where $M$ is a learnable parameter independent of the input, which acts as a memory of the whole training dataset. $A$ is the attention map inferred from \mtj{this learned dataset-level} prior knowledge; it is normalized in a similar way to self-attention (see Section 3.2). Finally, we update the input features from $M$ by the similarities in $A$. 

In practice, we use two different memory units $M_k$ and $M_v$ as the key and value, to increase the capability of the network. This slightly changes the computation of external attention to 
\begin{align}
    A & = \mathrm{Norm}(F M_k^T), \\
    F_{out} & = A M_v.
\end{align}

The computational complexity of external attention is ${O}(dSN)$; as $d$ and $S$ are hyper-parameters, the proposed algorithm is linear in the number of pixels. In fact, we find that a small $S$, e.g.\ 64, works well in  experiments. Thus, external attention is much more efficient than self-attention, allowing its direct application to large-scale inputs. \gmh{We also note that the computation load of external attention is roughly equivalent to a $1 \times 1$  convolution.}
\begin{figure}[t]
    \centering
    \includegraphics[width = \linewidth]{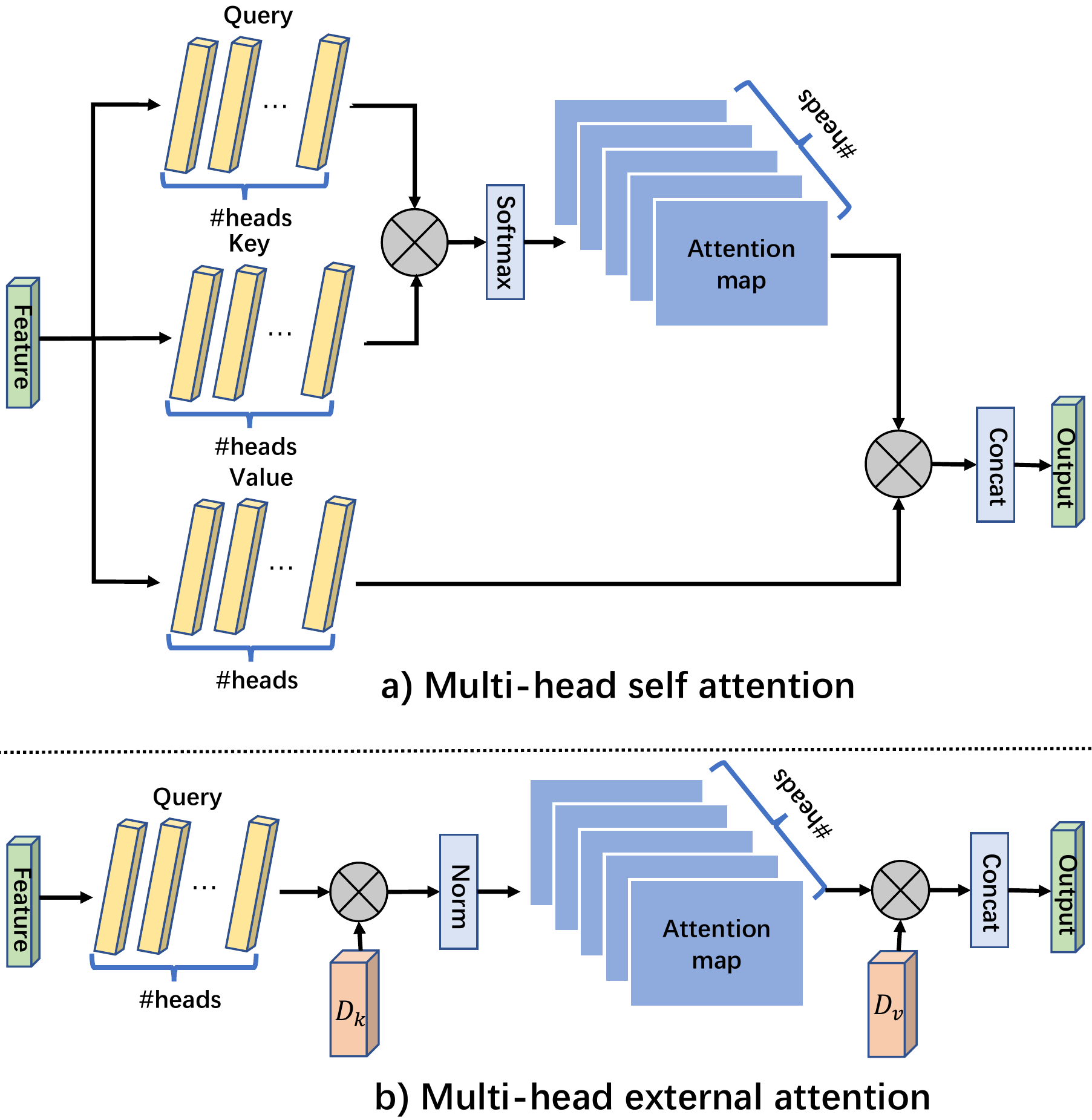}
    \caption{Multi-head self-attention and multi-head external-attention.}
    \label{fig:multi_head_attention}
\end{figure}

% \begin{figure}[t]
%     \centering
%     \includegraphics[width=\textwidth]{images/multi_head_attention.pdf}
%     \caption{
%     Multi-head self-attention versus multi-head external-attention.
%     }
%     \label{fig:seg_aritecture}
%     \vspace{-1ex}
% \end{figure}

\begin{figure*}[t]
    \centering
    \includegraphics[width=\textwidth]{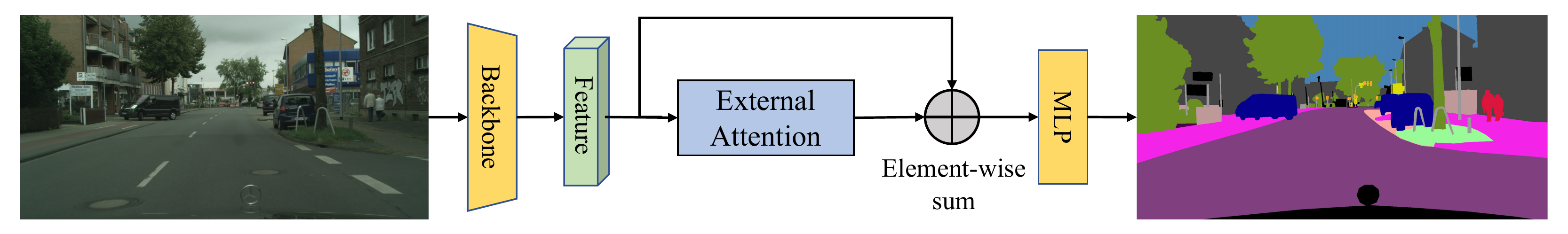}
    \caption{
     EANet architecture for semantic segmentation using our proposed external attention.
    }
    \label{fig:seg_aritecture}
    \vspace{-1ex}
\end{figure*}

% \begin{algorithm} \caption{Pseudo-code for multi-head external attention.} \label{alg.multi_head} 
% \begin{lstlisting}[language=Python]

\begin{algorithm} \caption{Pseudo-code for external attention.} \label{alg.single_head} 
\begin{lstlisting}[language=Python]
    # Input: F, an array with shape [B, N, C]  (batch size, pixels, channels)
    # Parameter: M_k, a linear layer without bias
    # Parameter: M_v, a linear layer without bias
    # Output: out, an array with shape [B, N, C]
    F = query_linear(F) # shape=(B, N, C)
    attn = M_k(F)       # shape=(B, N, M)
    attn = softmax(attn, dim=1)
    attn = l1_norm(attn, dim=2)
    out = M_v(attn)     # shape=(B, N, C)
\end{lstlisting}
\end{algorithm}

\subsection{Normalization}
\label{sec:novel_norm}

%Softmax is the most commonly used normalization method uniting with self-attention. However, we analyse the softmax normalization and find the softmax normalization is not the optimal way. We noticed that when the value of some feature vectors is particularly large. the relation between them and other feature vectors will be large which will cause the original meaning of attention map to be destroyed. It is the same when the value of some feature vectors is particularly small. We propose a new normalization way to solve this problem by using twice normalization with tiny computational requirements.

Softmax is employed in self-attention to normalize the attention map so that $\sum_{j} \alpha_{i,j} = 1$. However, the attention map is calculated by matrix multiplication. Unlike cosine similarity, the attention map is sensitive to the scale of the input features. To avoid this problem , we opt for the double-normalization proposed in~\cite{guo2021pct}, which separately normalizes  columns and  rows. This double-normalization is formulated as:
\begin{align}
\label{eq.norm}
    (\Tilde{\alpha})_{i,j} & = F M_k^T \\
    \hat{\alpha}_{i,j} & = {\exp{(\Tilde{\alpha}_{i,j})}}/{\sum\limits_{k}{\exp{(\Tilde{\alpha}_{k,j}})}} \\
    {\alpha}_{i,j} & = {\hat{\alpha}_{i,j}}/{\sum\limits_{k}\hat{\alpha}_{i,k}}
\end{align}

\mtj{A python-style pseudo-code for external attention is listed in Algorithm~\ref{alg.single_head}.}

\begin{algorithm} \caption{Pseudo-code for multi-head external attention.} \label{alg.multi_head} 
\begin{lstlisting}[language=Python]
    # Input: F, an array of shape [B, N, C_in]  (batch size, pixels, channels)
    # Parameter: M_k, a linear layer 
    # Parameter: M_v, a linear layer 
    # Parameter: H, number of heads
    # Output: out, an array of shape [B, N, C_in]
    F = query_linear(F) # shape=(B, N, C)
    F = F.view(B, N, H, C // H)
    F = F.permute(0, 2, 1, 3)
    attn = M_k(F)  # shape=(B, H, N, M)
    attn = softmax(attn, dim=2)
    attn = l1_norm(attn, dim=3)
    out = M_v(attn) # shape=(B, H, N, C // H)
    out = out.permute(0, 2, 1, 3)
    out = out.view(B, N, C) 
    out = W_o(out) # shape=(B, N, C_in)
\end{lstlisting}
\end{algorithm}

\subsection{Multi-head external attention}

\lzn{In Transformer~\cite{Vaswani2017attention}, self-attention is computed many times on different input channels, which is called multi-head attention.} \gmh{Multi-head attention can capture different relations between tokens, improving upon the capacity of single head attention.}  We use a similar approach for multi-head external attention as shown in Algorithm~\ref{alg.multi_head} and Fig.~\ref{fig:multi_head_attention}.

\gmh{Multi-head external attention can be written as:}
\begin{align}
\label{eq.multi_head}
    h_{i} & = \mathrm{ExternalAttention}(F_{i}, M_{k}, M_{v}), \\
    F_{out} & = \mathrm{MultiHead}(F, M_{k}, M_{v}) \\ 
        & = \mathrm{Concat}(h_{1}, \dots, h_{H}) W_{o}, 
    % F_{out} & = A M_v
\end{align}
\gmh{where $h_i$ is the $i$-th head, $H$ is the number of heads and $W_{o}$ is a linear transformation matrix making the dimensions of input and output consistent. $M_{k} \in \mathbb{R}^{S \times d}$ and $ M_{v} \in \mathbb{R}^{S \times d}$ are the shared memory units for different heads.}  

\gmh{The flexibility of this architecture allows us to balance between the number of head $H$ and number of \mtj{elements $S$ in} shared memory units. % $S$. 
For instance, we can multiply $H$ by $k$ while dividing $S$ by $k$.}

% \gmh{The different between multi-head external attention and multi-head self attention is that we share the memory units for different heads.}

\begin{table}[!t]
\centering
\caption{Ablation study on PASCAL VOC val set. Norm: Normalization method in attention. \#S: number of \mtj{elements in} memory units. OS: output stride of backbone. FCN~\cite{long2015fully}: fully convolutional network. SA: self-attention. EA: external-attention. DoubleNorm: normalization depicted as Eq.~\ref{eq.norm}}
\label{Tab.ablation}
\begin{tabular}{lrrrrr}
\toprule
\textbf{Method} & \textbf{Backbone} & \textbf{Norm} & \textbf{\#S} & \textbf{OS} & \textbf{mIoU(\%)} \\
\midrule

FCN & ResNet-50 & - & - & 16 & 75.7 \\ 
FCN + SA & ResNet-50 & Softmax & - & 16 & 76.2 \\ 
FCN + SA & ResNet-50 & DoubleNorm & - & 16 & 76.6 \\ 
FCN + EA & ResNet-50 & DoubleNorm & 8 & 16 & 77.1 \\ 
FCN + EA & ResNet-50 & DoubleNorm & 32 & 16 & 77.2 \\ 
FCN + EA & ResNet-50 & Softmax & 64 & 16 & 75.3 \\ 
FCN + EA & ResNet-50 & DoubleNorm & 64 & 16 & 77.4 \\ 
FCN + EA & ResNet-50 & DoubleNorm & 64 & 8 & 77.8 \\ 
FCN + EA & ResNet-50 & DoubleNorm & 256 & 16 & 77.0 \\ 
FCN + EA & ResNet-101 & DoubleNorm & 64 & 16 & \textbf{78.3} \\ 

\bottomrule
\end{tabular}
\end{table}

\begin{table*}
% \parbox{.45\linewidth}{
\centering
\caption{Experiments on ImageNet. Top1: top1 accuracy. EA: external-attention. MEA: multi-head external attention. EAMLP: proposed all MLP architecture. Failed: Unable to converge. EAMLP-BN: replace LN by BN in  T2T-ViT backbone’s MLP blocks(not external attention blocks).}
\label{Tab.ImageNet}
\begin{tabular}{lcccrrrrr}
\toprule
\textbf{Method} & \textbf{T2T-Transformer} & \textbf{T2T-Backbone} & \textbf{Input size} & \textbf{\#Heads} & \textbf{\#Memory units} & \textbf{\#Params(M)} & \textbf{\#Throughput}  & \textbf{Top1(\%)} \\
\midrule
T2T-ViT-7  & Performer  & Transformer & 224 x 224 & 1 & - & 4.3 & 2133.3 &  67.4  \\
T2T-ViT-7  & Performer  & Transformer & 224 x 224 & 4 & - & 4.3 & 2000.0 &   71.7  \\
T2T-ViT-14  & Performer  & Transformer & 224 x 224 & 6 & - & 21.5 & 969.7 &  81.5  \\
T2T-ViT-14  & Transformer  & Transformer & 224 x 224 & 6 & - & 21.5 & 800.0 &  81.7  \\
T2T-ViT-19  & Performer  & Transformer & 224 x 224 & 6 & - & 39.2 & 666.7 &  81.9  \\

\midrule

T2T-ViT-7 & EA & Transformer & 224 x 224 & 4 & - & 4.2 & 1777.8 &  71.7  \\
T2T-ViT-14 & EA & Transformer & 224 x 224 & 6 & - & 21.5 & 941.2 & 81.7 \\ 
\midrule

T2T-ViT-7  & Performer  & MEA & 224 x 224 & 1 & 256 & 4.3 & 2133.3  & 63.2  \\
T2T-ViT-7  & Performer  & MEA & 224 x 224 & 4 & 256 & 5.9  & 1684.2 & 66.8\\
T2T-ViT-7  & Performer  & MEA & 224 x 224 & 16 & 64 & 6.2 & 1684.2 & 68.6   \\
T2T-ViT-7  & Performer  & MEA & 384 x 384 & 16 & 64 & 6.3 &  615.4   & 70.9 \\
T2T-ViT-7  & Performer  & MEA & 224 x 224 & 16 & 128 & 6.3 & 1454.5 &  69.9  \\
T2T-ViT-7  & Performer  & MEA & 224 x 224 & 32 & 32 & 9.9 & 1280.0 & 70.5 \\
% T2T-ViT-14  & Performer  & MEA & 224 x 224 & 6 & 256 &  &   & 76.9 \\
T2T-ViT-14  & Performer  & MEA & 224 x 224 & 24 & 64 & 29.9 & 744.2 & 78.7  \\
% T2T-ViT-19  & Performer  & MEA & 224 x 224 & 6 & 256 & &    & 78.7 \\
T2T-ViT-19  & Performer  & MEA & 224 x 224 & 24 & 64 & 54.6 & 470.6 &  79.3   \\

\midrule
MLP-7 & MLP & MLP & 224 x 224 & - & - & &  & Failed   \\
EAMLP-7  & EA & MEA & 224 x 224 & 16 & 64 & 6.1 & 1523.8  & 68.9 \\
EAMLP-BN-7  & EA & MEA(BN) & 224 x 224 & 16 & 64 & 6.1 & 1523.8 & 70.0 \\
EAMLP-7  & EA & MEA & 384 x 384 & 16 & 64 & 6.2 & 542.4 &  71.7  \\
EAMLP-14  & EA & MEA & 224 x 224 & 24 & 64 & 29.9  & 711.1  & 78.9 \\
EAMLP-BN-14 & EA & MEA(BN) & 224 x 224 & 24 & 64 & &  & Failed   \\
EAMLP-19  & EA & MEA & 224 x 224 & 24 & 64 & 54.6 & 463.8 & 79.4 \\
EAMLP-BN-19 & EA & MEA(BN) & 224 x 224 & 24 & 64 & &  & Failed   \\

% \midrule
% T2T-ViT-7 & Our Attention  & Transformer & 4 & - &    71.7\% & \\
% T2T-ViT-14 & Our Attention & Transformer & 6 & - &  \textbf{81.7\%} & \\ 
\bottomrule
\end{tabular}
\end{table*}
% }
% \hfill

\begin{table}
% \parbox{.45\linewidth}{
\centering
\caption{Experiments on COCO object detection dataset. Results quoted are taken from~\cite{mmdetection}. Box AP: Box Average Precision.} %EA means External Attention}
\label{Tab.coco_detection}
\begin{tabular}{lcr}
\toprule
\textbf{Method} & \textbf{Backbone} & \textbf{Box AP} \\
\midrule
Faster RCNN~\cite{renNIPS15fasterrcnn} & ResNet-50 & 37.4 \\
Faster RCNN + 1EA & ResNet-50 & 38.5 \\
Mask RCNN~\cite{He_2017_mask} & ResNet-50 & 38.2 \\
Mask RCNN + 1EA & ResNet-50 & 39.0 \\
RetinaNet~\cite{retinanet} & ResNet-50 & 36.5 \\
RetinaNet + 1EA & ResNet-50 & 37.4 \\
Cascade RCNN~\cite{Cascade_rcnn} & ResNet-50 & 40.3 \\
Cascade RCNN + 1EA & ResNet-50 & 41.4 \\
Cascade Mask RCNN~\cite{Cascade_rcnn} & ResNet-50 & 41.2 \\
Cascade Mask RCNN + 1EA & ResNet-50 & 42.2 \\

\bottomrule
\end{tabular}
\end{table}

\begin{table}[!t]
% \parbox{.45\linewidth}{
\centering
\caption{Experiments on COCO instance segmentation dataset. Results quoted are taken from~\cite{mmdetection}. Mask AP: Mask Average Precision.}% EA means External Attention}
\label{Tab.coco_segmentation}
\begin{tabular}{lcr}
\toprule
\textbf{Method} & \textbf{Backbone} & \textbf{Mask AP} \\
\midrule
Mask RCNN~\cite{He_2017_mask} & ResNet-50 & 34.7	 \\
Mask RCNN + 1EA & ResNet-50 & 35.4 \\
Cascade RCNN~\cite{Cascade_rcnn} & ResNet-50 & 35.9 \\
Cascade RCNN + 1EA & ResNet-50 & 36.7 \\

\bottomrule
\end{tabular}
\end{table}

% \parbox{.45\linewidth}{
\begin{table}[!t]
\centering
\caption{Comparison to state-of-the-art methods on the PASCAL VOC test set w/o COCO pretraining.}
\label{Tab.VOC}
\begin{tabular}{lcr}
\toprule
\textbf{Method} & \textbf{Backbone} & \textbf{mIoU(\%)} \\
\midrule
\specialrule{0em}{0pt}{1pt}
PSPNet~\cite{zhao2017pspnet} & ResNet-101  &  82.6  \\
DFN~\cite{DFN} & ResNet-101  &  82.7  \\
EncNet~\cite{EncNet} & ResNet-101 &  82.9  \\
SANet~\cite{SANet} & ResNet-101 &  83.2  \\
DANet~\cite{fu2019dual} & ResNet-101 &  82.6  \\
CFNet~\cite{CFNet} & ResNet-101 &  84.2  \\
SpyGR~\cite{spyGR} & ResNet-101 & 84.2 \\
\midrule
EANet (Ours) & ResNet-101 &  84.0 \\
\bottomrule
\end{tabular}
% }
\end{table}

\begin{table}[!t]
\centering
\caption{Comparison to state-of-the-art methods on the ADE20K val set.}
\label{Tab.ADE}
\begin{tabular}{lcr}
\toprule
\textbf{Method} & \textbf{Backbone} & \textbf{mIoU(\%)} \\
\midrule
PSPNet~\cite{zhao2017pspnet}  & ResNet-101 &  43.29  \\
PSPNet~\cite{zhao2017pspnet}  & ResNet-152 &  43.51  \\
PSANet~\cite{zhao2018psanet}  & ResNet-101 &  43.77  \\
EncNet~\cite{EncNet} & ResNet-101  &  44.65  \\
CFNet~\cite{CFNet}  & ResNet-101  &  44.89  \\
PSPNet~\cite{zhao2017pspnet}  & ResNet-269 &  44.94  \\
OCNet~\cite{yuan2019ocnet}  & ResNet-101  & 45.04 \\
ANN~\cite{Zhu_2019_ANN} & ResNet-101  & 45.24 \\
DANet~\cite{fu2019dual}  & ResNet-101  &  45.26  \\
OCRNet~\cite{yuan2020objectcontextual} & ResNet-101  &  45.28  \\
\midrule
EANet (Ours) & ResNet-101 &  \textbf{45.33} \\
\bottomrule
\end{tabular}
\end{table}

\begin{table}[!t]
\centering
\caption{Comparison to state-of-the-art methods on the cityscapes val set; results quoted are taken from~\cite{mmseg2020}. }
\label{Tab.city}
\begin{tabular}{lcr}
\toprule
\textbf{Method} & \textbf{Backbone} \textbf{mIoU(\%)} \\
\midrule
EncNet~\cite{EncNet} & ResNet-101 &  78.7  \\
APCNet~\cite{He_2019_APC} & ResNet-101 &  79.9  \\
ANN~\cite{Zhu_2019_ANN} & ResNet-101 & 80.3 \\
DMNet~\cite{DMNet_He} & ResNet-101 & 80.7 \\
GCNet~\cite{cao2019gcnet}& ResNet-101 & 80.7 \\
PSANet~\cite{zhao2018psanet} & ResNet-101 &  80.9	  \\
EMANet~\cite{li19} & ResNet-101 &  81.0	  \\
PSPNet~\cite{zhao2017pspnet} & ResNet-101 &  81.0  \\
DANet~\cite{fu2019dual} & ResNet-101 &  82.0  \\
\midrule
EANet (Ours) & ResNet-101 &  81.7 \\
\bottomrule
\end{tabular}
% }
\end{table}

\begin{table}[!t]
% \parbox{.45\linewidth}{
\centering
\tabcaption{Comparison to GAN methods on cifar-10 dataset.}
\label{Tab.gan_cifar}
\begin{tabular}{lrr}
    \toprule
    \textbf{Method} & \textbf{FID} & \textbf{IS} \\
    \midrule
    DCGAN~\cite{dcgan}  & 49.030  &  6.638  \\
    LSGAN~\cite{LSGAN}  & 66.686  &  5.577  \\
    WGAN-GP~\cite{wgan_gp}  & 25.852  & 7.458  \\
    ProjGAN~\cite{projGAN}  & 33.830  & 7.539  \\
    SAGAN~\cite{SANet}  & 14.498  & 8.626  \\
    \midrule
    EAGAN (Ours) & \textbf{14.105} & \textbf{8.630} \\
\bottomrule
\end{tabular}
\end{table}

\begin{table}[!t]
% \parbox{.45\linewidth}{
\centering
\tabcaption{Comparison to GAN methods on Tiny-ImageNet dataset.}
\label{Tab.gan_tiny_imagenet}
\begin{tabular}{lrr}
    \toprule
    \textbf{Method} & \textbf{FID} & \textbf{IS} \\
    \midrule
    DCGAN~\cite{dcgan}  & 91.625  &  5.640  \\
    LSGAN~\cite{LSGAN}  & 90.008  &  5.381  \\
    ProjGAN~\cite{projGAN}  & 89.175  & 6.224  \\
    SAGAN~\cite{SANet}  & 51.414  & 8.342  \\
    \midrule
    EAGAN (Ours) & \textbf{48.374} & \textbf{8.673} \\
\bottomrule
\end{tabular}
\end{table}

\begin{figure*}[t]
    \centering
    \includegraphics[width=.75\linewidth]{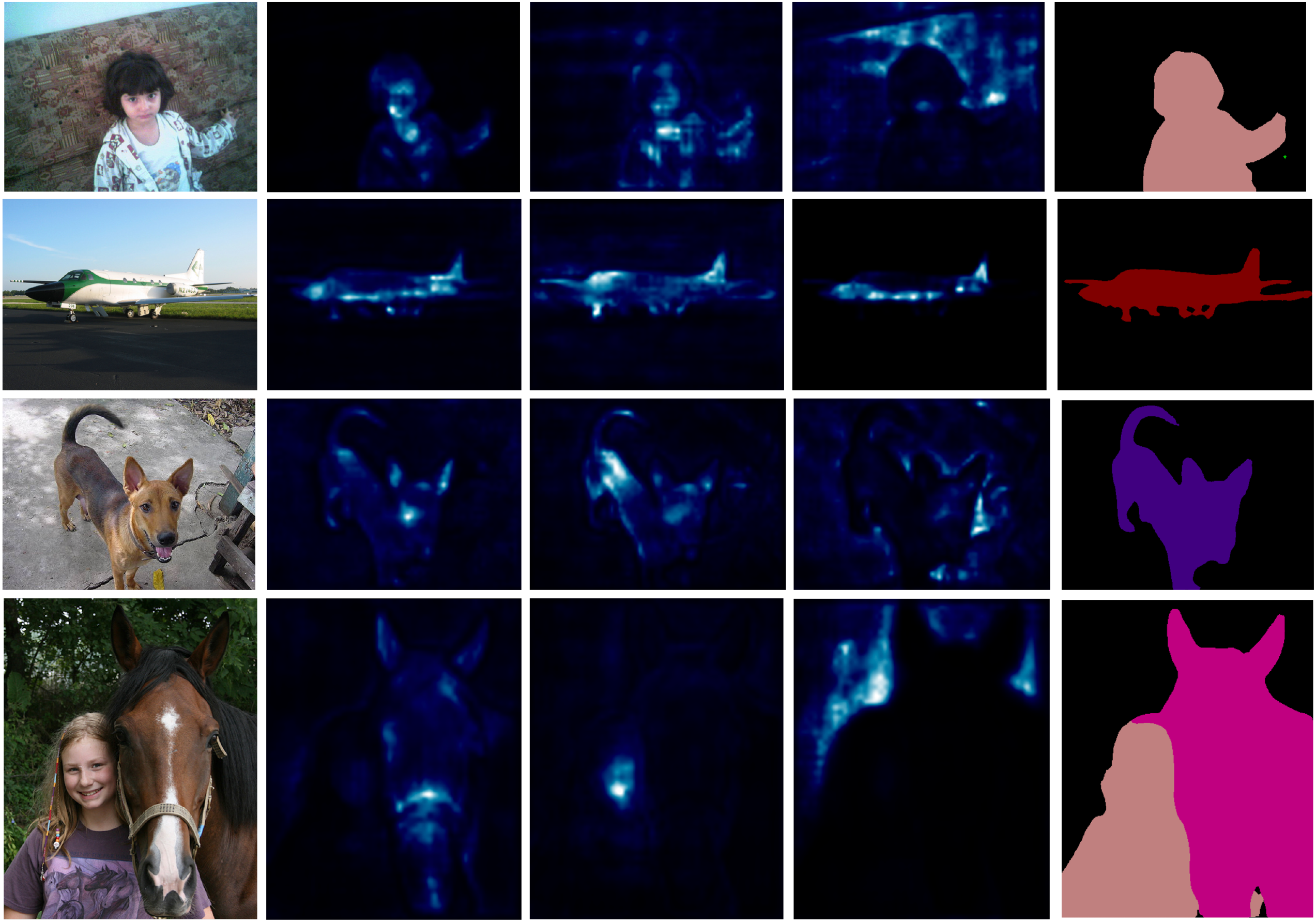}
    \caption{
   Attention map and segmentation results on Pascal VOC test set. Left to right: input images, attention maps w.r.t. three selected entries in the external memory, segmentation results.
    }
    \label{fig:seg_attention}
\end{figure*}

\section{Experiments}

\begin{figure*}[t]
    \centering
    \includegraphics[width=\textwidth]{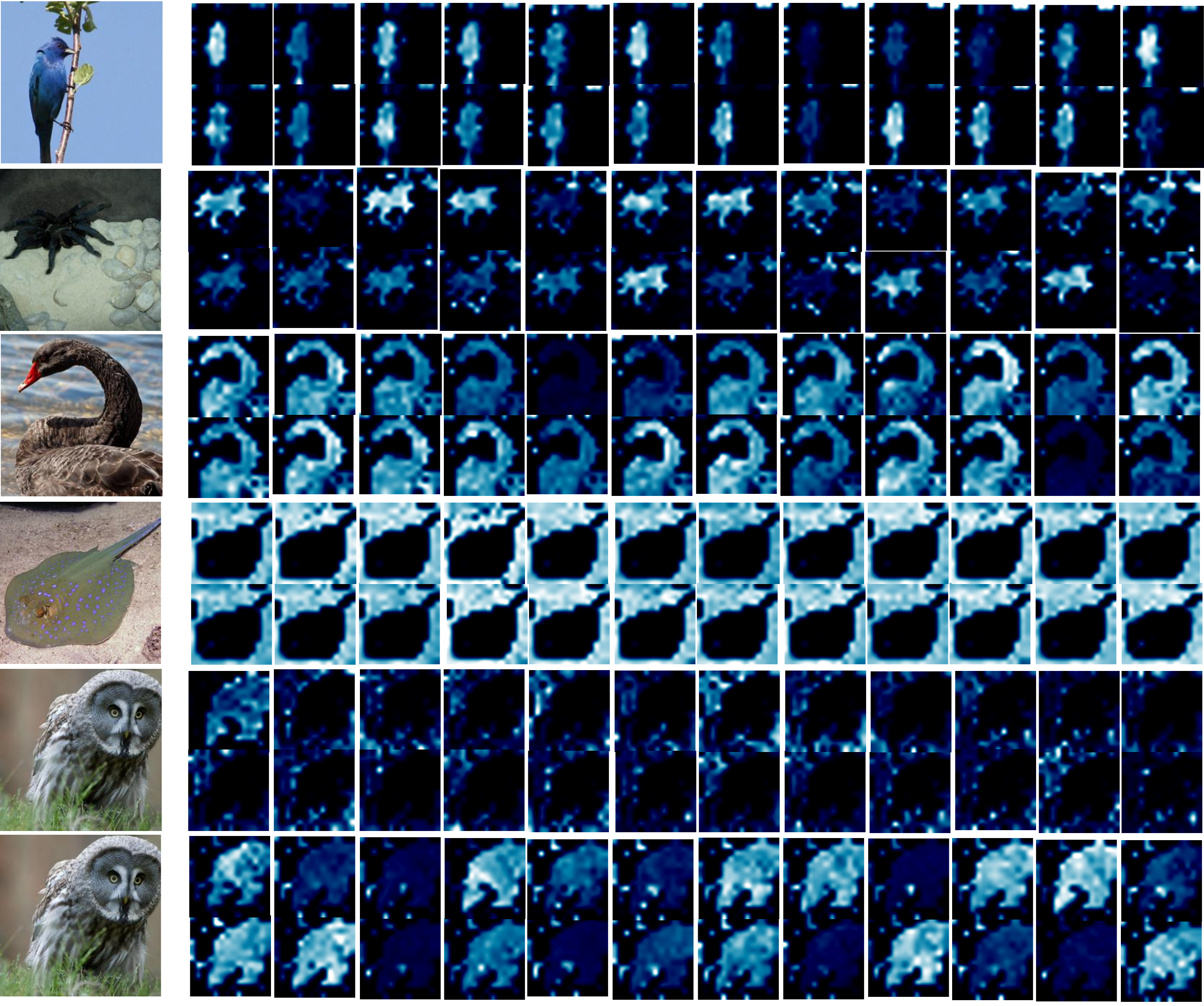}
    \caption{
   Multi-head attention map in the last layer of EAMLP-14 on ImageNet val set. Left: Input image  Others: 24 head attention map in the last layer of EAMLP-14 for the ImageNet val set. Last two rows: attention of two different rows of $M_k$ to the image patches.
    }
    \label{fig:cls_attention}
\end{figure*}

\begin{figure}[t]
    \centering
    \includegraphics[width = \linewidth]{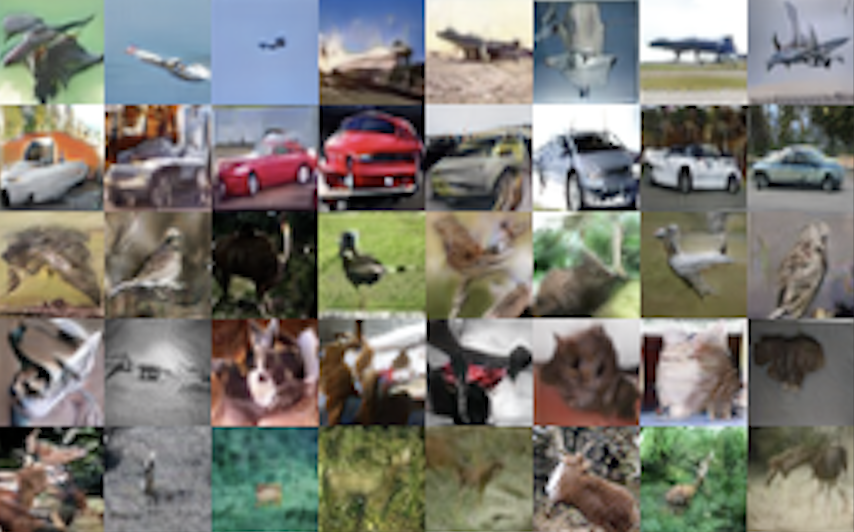}
    \caption{Images generated using our method on cifar-10.}
    \label{fig:imggen}
\end{figure}

\begin{table*}[t!]
    \centering
    % \small
    \caption{Comparison using the ShaperNet part segmentation dataset. pIoU: part-average intersection-over-union. Results quoted are taken from  cited papers.}
    \setlength{\tabcolsep}{1pt}
    \begin{tabular}{l|r|rrrrrrrrrrrrrrrr}
        \toprule
        \textbf{Method}                         & {pIoU}        & \tabincell{c}{{air}-\\{plane}} & {bag}         & {cap}         & {car}         & {chair}       & \tabincell{c}{{ear}-\\{phone}} & {guitar}      & {knife}       & {lamp}        & {laptop}      & \tabincell{c}{motor-                                                                                 \\bike} & {mug} & {pistol} & {rocket} & \tabincell{c}{{skate}-\\{board}}& {table} \\
        \midrule
        PointNet~\cite{qi2016pointnet}          & 83.7          & 83.4                                                 & 78.7          & 82.5          & 74.9          & 89.6          & 73.0                                                 & 91.5          & 85.9          & 80.8          & 95.3          & 65.2                 & 93.0          & 81.2          & 57.9          & 72.8          & 80.6          \\
        Kd-Net~\cite{Klokov2017kdnet}           & 82.3          & 80.1                                                 & 74.6          & 74.3          & 70.3          & 88.6          & 73.5                                                 & 90.2          & 87.2          & 81.0          & 94.9          & 57.4                 & 86.7          & 78.1          & 51.8          & 69.9          & 80.3          \\

        SO-Net~\cite{li2018sonet}               & 84.9          & 82.8                                                 & 77.8          & 88.0          & 77.3          & 90.6          & 73.5                                                 & 90.7          & 83.9          & 82.8          & 94.8          & 69.1                 & 94.2          & 80.9          & 53.1          & 72.9          & 83.0          \\
        PointNet++~\cite{qi2017pointnet++}      & 85.1          & 82.4                                                 & 79.0          & 87.7          & 77.3          & 90.8          & 71.8                                                 & 91.0          & 85.9          & 83.7          & 95.3          & 71.6                 & 94.1          & 81.3          & 58.7          & 76.4          & 82.6          \\
        PCNN~\cite{Atzmon2018point}             & 85.1          & 82.4                                                 & 80.1          & 85.5          & 79.5          & 90.8          & 73.2                                                 & 91.3          & 86.0          & 85.0          & 95.7          & 73.2                 & 94.8          & 83.3          & 51.0          & 75.0          & 81.8          \\

        DGCNN~\cite{wang2019dynamic}            & 85.2          & 84.0                                                 & 83.4          & 86.7          & 77.8          & 90.6          & 74.7                                                 & 91.2          & 87.5          & 82.8          & 95.7          & 66.3                 & 94.9          & 81.1          & 63.5          & 74.5          & 82.6          \\
        P2Sequence~\cite{liu2019Point2Sequence} & 85.2          & 82.6                                                 & 81.8          & 87.5          & 77.3          & 90.8          & 77.1                                                 & 91.1          & 86.9          & 83.9          & 95.7          & 70.8                 & 94.6          & 79.3          & 58.1          & 75.2          & 82.8          \\
        PointConv~\cite{Wu2019pointconv}        & 85.7          & -                                                    & -             & -             & -             & -             & -                                                    & -             & -             & -             & -             & -                    & -             & -             & -             & -             & -             \\
        PointCNN~\cite{li2018pointcnn}          & 86.1          & 84.1                                                 & \textbf{86.5} & 86.0          & 80.8          & 90.6          & 79.7                                                 & \textbf{92.3} & 88.4 & 85.3          & 96.1 & \textbf{77.2}        & 95.2          & 84.2 & \textbf{64.2} & \textbf{80.0} & 83.0          \\
        PointASNL~\cite{yan2020pointasnl}       & 86.1          & 84.1                                                 & 84.7          & 87.9          & 79.7          & \textbf{92.2} & 73.7                                                 & 91.0          & 87.2          & 84.2          & 95.8          & 74.4                 & 95.2          & 81.0          & 63.0          & 76.3          & 83.2          \\
        RS-CNN~\cite{Liu2019rscnn}              & 86.2          & 83.5                                                 & 84.8          & 88.8          & 79.6          & 91.2          & \textbf{81.1}                                        & 91.6          & 88.4 & 86.0          & 96.0          & 73.7                 & 94.1          & 83.4          & 60.5          & 77.7          & 83.6          \\
        PCT~\cite{guo2021pct}                            & 86.4 & 85.0                                        & 82.4          & 89.0 & 81.2 & 91.9          & 71.5                                                 & 91.3          & 88.1          & \textbf{86.3} & 95.8          & 64.6                 & \textbf{95.8} & 83.6          & 62.2          & 77.6          & \textbf{83.7} \\
        \midrule
        \textbf{Ours}                            & \textbf{86.5} & \textbf{85.1}                                        & 85.7          & \textbf{90.3} & \textbf{81.6} & 91.4          & 75.9                                                 & 92.1          & \textbf{88.7}          & 85.7 & \textbf{96.2}       & 74.8                 & 95.7 & \textbf{84.3}          & 60.2         & 76.2          & 83.5 \\

        \bottomrule
    \end{tabular}
    \label{Tab.ShapeNet}
\end{table*}

\begin{table}[t!]
%\parbox{.52\linewidth}{
    \centering
    \caption{Comparison to state-of-the-art methods on ModelNet40 classification dataset. Accuracy: overall accuracy. All results quoted are taken from the cited papers. P = points, N = normals.}
    \begin{tabular}{l|lrr}
        \toprule
        \textbf{Method}                         & \textbf{input} & \textbf{\#points} & \textbf{Accuracy} \\
        \midrule
        PointNet~\cite{qi2016pointnet}          & P              & 1k                & 89.2\%            \\
        A-SCN~\cite{Xie_2018_CVPR} & P & 1k & 89.8\%            \\
        
        SO-Net~\cite{li2018sonet}               & P, N           & 2k                & 90.9\%            \\
        Kd-Net~\cite{Klokov2017kdnet}           & P              & 32k               & 91.8\%            \\
        PointNet++~\cite{qi2017pointnet++}      & P              & 1k                & 90.7\%            \\
        PointNet++~\cite{qi2017pointnet++}      & P, N           & 5k                & 91.9\%            \\
        PointGrid~\cite{Le2018Pointgrid}        & P              & 1k                & 92.0\%            \\
        PCNN~\cite{Atzmon2018point}             & P              & 1k                & 92.3\%            \\
        PointWeb~\cite{Zhao2019pointweb}        & P              & 1k                & 92.3\%            \\
        PointCNN~\cite{li2018pointcnn}          & P              & 1k                & 92.5\%            \\
        PointConv~\cite{Wu2019pointconv}        & P, N           & 1k                & 92.5\%            \\
        A-CNN~\cite{Komarichev2019acnn}         & P, N           & 1k                & 92.6\%            \\
        P2Sequence~\cite{liu2019Point2Sequence} & P              & 1k                & 92.6\%            \\
        KPConv~\cite{Thomas2019kpconv}          & P              & 7k                & 92.9\%            \\
        DGCNN~\cite{wang2019dynamic}            & P              & 1k                & 92.9\%            \\
        RS-CNN~\cite{Liu2019rscnn}              & P              & 1k                & 92.9\%            \\
        PointASNL~\cite{yan2020pointasnl}       & P              & 1k                & 92.9\%            \\
        PCT ~\cite{guo2021pct}                  & P              & 1k                & 93.2\%   \\
        \midrule
        EAT (Ours)         & P              & 1k                & \textbf{93.4\%}   \\
        \bottomrule
    \end{tabular}
    \label{Tab.ModelNet40.classification}
%}
\end{table}

\begin{table*}
	\centering
	\caption{Computational requirements compared to self-attention and its variants. MACs: Multiply-accumulate operations.}
\setlength{\tabcolsep}{1.5pt}
\label{Tab.Computational}
\begin{tabular}{lcccccccc}
\toprule
\textbf{Method} & SA~\cite{Vaswani2017attention} & DA~\cite{fu2019dual} & $A^2$~\cite{DBLP:journals/corr/abs-1810-11579} & APC~\cite{He_2019_APC} & DM~\cite{DBLP:conf/iccv/HeD019} & ACF~\cite{DBLP:conf/cvpr/0005ZWX19} & Ham~\cite{geng2021is} & EA (ours) \\
\midrule
\textbf{Params} & 1.00M & 4.82M & 1.01M & 2.03M & 3.00M & 0.75M & \textbf{0.50M} & 0.55M \\
\textbf{MACs} & 292G & 79.5G & 25.7G & 17.6G & 35.1G & 79.5G & 17.6G & \textbf{9.2G}\\
\bottomrule
\end{tabular}
\end{table*}

% !TEX root = ../main.tex

We have conducted experiments on image classification, \gmh{object detection}, semantic segmentation, \gmh{instance segmentation}, image generation, point cloud classification, and point cloud segmentation tasks to assess the effectiveness of our proposed external attention approach. All experiments were implemented with Jittor~\cite{hu2020jittor} and\mtj{/or} Pytorch~\cite{paszke2019pytorch} deep learning frameworks.

\subsection{Ablation study}
\gmh{\mtj{To validate the proposed modules in our full model, we} %Ablation study is 
conducted experiments on the PASCAL VOC segmentation dataset~\cite{DBLP:journals/ijcv/EveringhamGWWZ10}. Fig.~\ref{fig:seg_aritecture} depicts the architecture used for ablation study, which takes the FCN~\cite{long2015fully} as the feature backbone. The batch size and total number of iterations were set to 12 and 30,000 respectively. We focus on the number of memory units, self attention versus external attention, the backbone, the normalization method, and output stride of the backbone. As shown in Table~\ref{Tab.ablation}, we can observe external attention provides better accuracy than self attention on the Pascal VOC dataset. Choosing a suitable number of memory units is important to quality of results. The normalization method can produce a huge positive effect on external attention and make a improvement on self-attention.}

\subsection{Visual analysis}

\gmh{Attention maps using external attention for segmentation (see Fig.~\ref{fig:seg_aritecture}) and multi-head external attention for classification (see Section~\ref{sec:exp:imgclf}) are shown in Figs.~\ref{fig:seg_attention} and~\ref{fig:cls_attention}, respectively. We randomly select a row $M^{i}_{k}$ from a memory unit $M_{k}$ in a layer. Then the attention maps are depicted by calculating the attention of $M^{i}_{k}$ to the input feature. We observe that the learned attention maps focus on meaningful objects or background \mtj{for segmentation task} as in Fig.~\ref{fig:seg_attention}. The last two rows in Fig.~\ref{fig:cls_attention} suggest that different rows of $M_{k}$ pay attention to different regions. Each head of multi-head external attention can activate regions of interest to different extents, as shown in Fig~\ref{fig:cls_attention}, improving the representation ability of external attention.}

\subsection{Image classification}
\label{sec:exp:imgclf}

ImageNet-1K\cite{imagenet_deng} is a widely-used dataset for image classification. We replaced the Performer~\cite{performer2021rethinking} and multi-head self-attention blocks in T2T-ViT~\cite{yuan2021tokenstotoken} with  external attention and multi-head external attention.
For fairness, other hyperparameter settings were the same as T2T-ViT. Experimental results in Table~\ref{Tab.ImageNet} show that external attention achieves a better result than \mtj{Performer~\cite{performer2021rethinking}} and a about 2\% point lower results than multi-head attention. We find multi-head mechanism is necessary to both self-attention and external attention. We also attempt the strategy proposed by MoCo V3~\cite{chen2021empirical} to replace LayerNorm(LN)~\cite{ba2016layer} by BatchNorm(BN)~\cite{ioffe2015batch} in the T2T-ViT backbone’s MLP blocks(not external attention blocks).  We observe a 1\% improvement on our EAMLP-7. However, it produce failed cases in our big model EAMLP-14 and EAMLP-19.

% comparable results to multi-head attention and performer~\cite{performer2021rethinking}.

\subsection{Object detection and instance segmentation}
\gmh{The MS COCO dataset~\cite{lin2015coco} is a popular benchmark for object detection and instance segmentation.  It contains more than 200,000 images with over 500,000 annotated object instances from 80 categories.}

\gmh{MMDetection~\cite{mmdetection} is a widely-used toolkit for object detection  and instance segmentation. We conducted our object detection and instance segmentation experiments using MMDetection with a RestNet-50 backbone, applied to the COCO dataset. We only added our external attention at the end of Resnet stage 4. Results in Tabs.~\ref{Tab.coco_detection} and~\ref{Tab.coco_segmentation} show that external attention brings about 1\% improvement in accuracy for both object detection and instance segmentation tasks. }

\subsection{Semantic segmentation}
In this experiment, we adopt the semantic segmentation architecture in Fig.~\ref{fig:seg_aritecture}, referring to it as EANet, and applied it to the Pascal VOC~\cite{DBLP:journals/ijcv/EveringhamGWWZ10}, ADE20K~\cite{DBLP:journals/ijcv/ZhouZPXFBT19} and cityscapes~\cite{cordts2016cityscapes} datasets. 

Pascal VOC contains 10,582 images for training, 1,449 images for validation and 1,456 images for testing. It has 20 foreground object classes and a background class for segmentation. We used dilated ResNet-101 with an output stride of 8 as the backbone, as for all compared methods; it was 
pre-trained on ImageNet-1K. A poly-learning rate policy was adopted during training. The initial learning rate, batch size and input size were set to 0.009, 16 and $513\times 513$. We first trained for 45k iterations on the training set and then fine-tuned for 15k iterations on the trainval set. Finally we used multi-scale and flip tests on the test set. 
Visual results are shown in Fig.~\ref{fig:seg_attention} and quantitative results are given in Table~\ref{Tab.VOC}\mtj{: our method can achieve comparable performance to the state-of-the-art methods}.

ADE20K is a more challenging dataset with 150 classes, and 20K, 2K, and 3K images for training, validation, and testing, respectively.  We adopted dilated ResNet-101 with an output stride of 8 as the backbone. The experimental configuration was the same as for mmsegmentation~\cite{mmseg2020}, training ADE20K for 160k iterations.
Results in Table~\ref{Tab.ADE} show that our method outperforms others on the ADE20K val set.

{Cityscapes} contains 5,000 high quality pixel-level finely annotated labels in 19 semantic classes for urban scene understanding. Each image is $1024 \times 2048$ pixels.  It is divided into 2975, 500 and 1525 images for training, validation and testing. (It also contains 20,000 coarsely annotated images, which we did not use in our experiments). We adopted dilated ResNet-101 with an output stride of  8 as the backbone for all  methods. The experimental configurations was again the same as for mmsegmentation, training cityscapes with 80k iterations. 
Results in Table~\ref{Tab.city} show that our method achieves comparable results to \mtj{state-of-the-art method, i.e, DANet~\cite{fu2019dual},} %other methods 
on the cityscapes val set.

% % %   \begin{minipage}[c]{0.485\linewidth}
%     \centering
%     \tabcaption{Compared with some GAN methods on cifar-10 dataset.}
%     \label{Tab.gan}
%     \begin{tabular}{lcc}
%     \toprule
%     \textbf{Method} & \textbf{FID} & \textbf{IS} \\
%     \midrule
%     DCGAN~\cite{dcgan}  & 49.03  &  6.638  \\
%     LSGAN~\cite{LSGAN}  & 66.686  &  5.577  \\
%     WGAN-GP~\cite{wgan_gp}  & 25.852  & 7.458  \\
%     ProjGAN~\cite{projGAN}  & 33.830  & 7.539  \\
%     SAGAN~\cite{SANet}  & 14.498  & 8.626  \\
%     \midrule
%     EAGAN(Ours) & \textbf{14.105} & \textbf{8.630} \\
%     \bottomrule
%     \end{tabular}
% %   \end{minipage}\hfill
%   \begin{minipage}[c]{0.485\linewidth}
%     \centering
%     \includegraphics[width = \linewidth]{images/cifar_generation.png}
%     \figcaption{Generated images using our method on cifar-10.}
%     \label{fig:imggen}
%   \end{minipage}
% \end{figtab}

\subsection{Image generation}

Self-attention is commonly used in  image generation, a representative approach being SAGAN~\cite{Zhang2019sagan}. We replaced the self-attention mechanism in SAGAN by our external attention approach in both the generator and discriminator to obtain our EAGAN model. All experiments were based on the popular PyTorch-StudioGAN repo~\cite{kang2020ContraGAN}. The hyper-parameters use the default configuration for SAGAN. We used Frechet Inception Distance (FID)~\cite{heusel2018gans_fid} and Inception Score (IS)~\cite{salimans2016improved} as our evaluation metric. Some generated images are shown in Fig.~\ref{fig:imggen} and quantitative results are given in Tabs.~\ref{Tab.gan_cifar} and~\ref{Tab.gan_tiny_imagenet}: external attention provides better results than SAGAN and some other GANs.

\subsection{Point cloud classification}
ModelNet40~\cite{DBLP:conf/cvpr/WuSKYZTX15} is a popular benchmark for 3D shape classification, containing 12,311 CAD models in 40 categories. It has 9,843 training samples and 2,468 test samples. Our EAT model replaces all self-attention modules in PCT~\cite{guo2021pct}. We sampled 1024 points on each shape and augmented the input with random translation, anisotropic scaling, and dropout, following PCT~\cite{guo2021pct}. Table~\ref{Tab.ModelNet40.classification} indicates that our method outperforms all others, including other attention-based methods like PCT. Our proposed method provides an outstanding backbone for both 2D and 3D vision.

\subsection{Point cloud segmentation}
We conducted a point cloud segmentation experiment on the ShapeNet part dataset~\cite{DBLP:journals/tog/YiKCSYSLHSG16}. It has 14,006 3D models in the training set and 2,874 in the evaluation set. Each shape is segmented into parts, with 16 object categories and 50 part labels in total. We  followed the experimental setting in PCT~\cite{guo2021pct}. EAT 
achieved the best results on this dataset, as indicated in Table~\ref{Tab.ShapeNet}.

\subsection{Computational requirements}

The linear complexity with respect to the size of the input brings about a significant advantage in efficiency. We  compared external attention (EA) \mtj{module} to standard self-attention (SA)~\cite{Vaswani2017attention} and several of its variants in terms of numbers of parameters and inference operations for an input size of $1\times512\times128\times128$, 
giving the results in Table~\ref{Tab.Computational}. External attention requires only half of the parameters needed by self-attention and  is 32 times faster. Compared to the best variant, external attention is still about twice as fast.

% !TEX root = ../main.tex
\section{Conclusions}
This paper has presented external attention, a novel lightweight yet effective attention mechanism useful for various visual tasks.
The two external memory units adopted in external attention can be viewed as  dictionaries for the whole dataset and are capable of learning  more representative features for the input while reducing  computational cost. 
We hope external attention will inspire practical applications and research into its use in other domains such as NLP.

% or


\ifCLASSOPTIONcompsoc
  \section*{Acknowledgments}
\else
  \section*{Acknowledgment}
\fi

\gmh{This work was supported by the Natural Science Foundation of China (Project 61521002). We would like to thank Xiang-Li Li for his kind help in experiments, Jun-Xiong Cai for helpful discussions, and  Prof. Ralph R. Martin for his insightful suggestions and great help in writing.}

\ifCLASSOPTIONcaptionsoff
  \newpage
\fi

\bibliographystyle{IEEEtran}
\bibliography{egbib}

\begin{IEEEbiography}[{\includegraphics[width=1in,height=1.25in,clip,keepaspectratio]{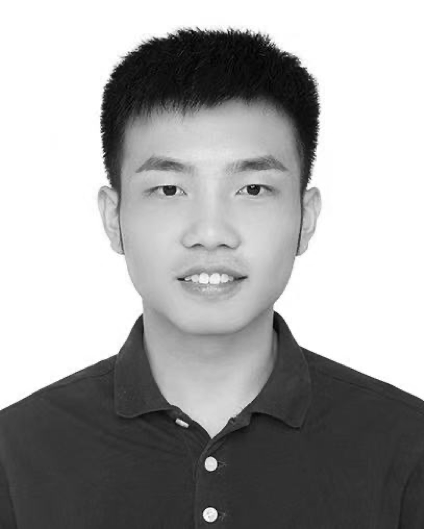}}]{Meng-Hao Guo}
received his bachelor's degree from Xidian University. Now he is a Ph.D. candidate supervised by Prof. Shi-Min Hu in the Department of Computer Science and Technology at Tsinghua University, Beijing, China. His research interests include computer vision, computer graphics,  and machine learning.
\end{IEEEbiography}

\begin{IEEEbiography}[{\includegraphics[width=1in,height=1.25in,clip,keepaspectratio]{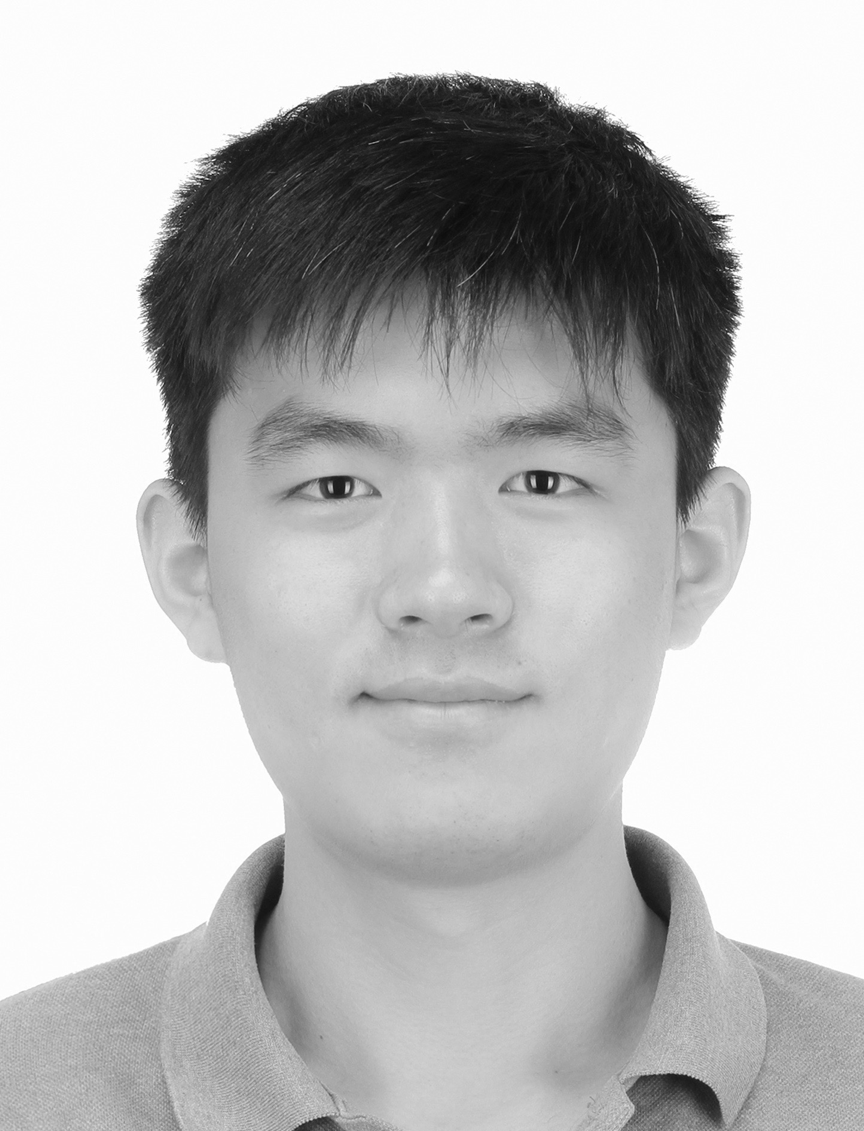}}]{Zheng-Ning Liu}
received his bachelor's degree in computer science from Tsinghua University in 2017. He is currently a Ph.D. candidate in  Computer Science at Tsinghua University. His research interests include 3D computer vision, 3D reconstruction, and computer graphics.
\end{IEEEbiography}

\begin{IEEEbiography}[{\includegraphics[width=1in,height=1.25in,clip,keepaspectratio]{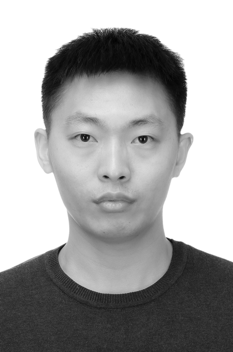}}]{Tai-Jiang Mu}
is currently an assistant researcher at Tsinghua University, where he received his B.S. and Ph.D. degrees in Computer Science in 2011 and 2016, respectively. His research interests include computer vision, robotics and computer graphics.
\end{IEEEbiography}

\begin{IEEEbiography}[{\includegraphics[width=1in,height=1.25in,clip,keepaspectratio]{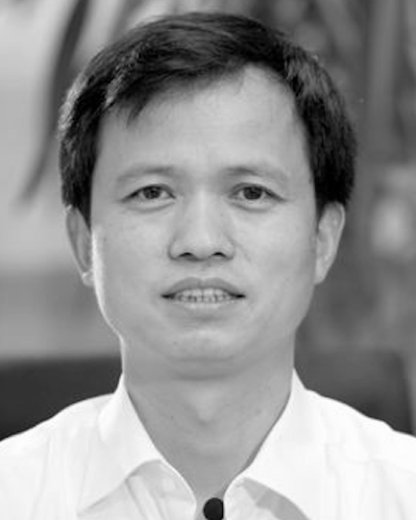}}]{Shi-Min Hu}
is currently a professor in Computer Science at Tsinghua University. He received a Ph.D. degree from Zhejiang University in 1996. His research interests include geometry processing, image \& video processing, rendering, computer animation, and CAD. He has published more than 100 papers in journals and refereed conferences. He is Editor-in-Chief of Computational Visual Media, and on the editorial boards of several journals, including Computer Aided Design and Computer \& Graphics.
\end{IEEEbiography}

% \begin{IEEEbiographynophoto}{Menghao.jpeg}
% % Biography text here.

% % insert where needed to balance the two columns on the last page with
% %\newpage



%\vfill


\end{document}